\documentclass{article} 
\usepackage{arxiv}

\usepackage[english]{babel}

\usepackage{tikz}
\usepackage{graphicx}
\usepackage{amsfonts} 
\usepackage{amsmath}
\usepackage{subfigure}
\usepackage{algorithmic}
\usepackage{subcaption}
\usepackage{multirow}
\usepackage{multicol}
\usepackage{wrapfig}
\usepackage{url}
\DeclareMathOperator*{\argmin}{arg\,min}
\usepackage[lined, ruled, linesnumbered, noend, scleft, nofillcomment]{algorithm2e}

\def\orcidID#1{\unskip$^{[#1]}$} 

\def\institute#1{\gdef\@institute{#1}}

\title{Learning From Scenarios for Repairable Stochastic Scheduling}
\author{Kim van den Houten \orcidID{0009-0001-7540-980X} \and
David M.J. Tax\orcidID{0000-0002-5153-9087} \and Esteban Freydell\orcidID{0009-0001-6992-2869} \and
Mathijs de Weerdt\orcidID{0000-0002-0470-6241}}

\begin{document}
\maketitle
\begin{abstract}
 When optimizing problems with uncertain parameter values in a linear objective, decision-focused learning enables end-to-end learning of these values.  We are interested in a stochastic scheduling problem, in which processing times are uncertain, which brings uncertain values in the constraints, and thus repair of an initial schedule may be needed. Historical realizations of the stochastic processing times are available. We show how existing decision-focused learning techniques based on stochastic smoothing can be adapted to this scheduling problem. We include an extensive experimental evaluation to investigate in which situations decision-focused learning outperforms the state of the art, i.e., scenario-based stochastic optimization.
\end{abstract}

\section{Introduction}
Decision-making can be challenging due to the stochastic nature of real-world processes. This complexity is evident in various contexts, such as manufacturing, where uncertain processing times make it challenging to meet strict customer deadlines. Formulating Constrained Optimization (CO) models for these problems is common, but unknown parameter values during decision-making add challenges, because wrong estimates of the parameters can lead to infeasibilities. 

In practice, such infeasibilities are repaired when reality unfolds.  For instance, in a manufacturing system, tasks may be postponed due to delays in earlier stages to maintain the factory's flow.  Various repair policies and schedule definitions are used across different contexts. 

Historical data, represented as scenarios of unknown parameters like task duration, are often available. Simple averaging of these scenarios is a common yet naive approach that ignores uncertainty. Stochastic programming \cite{shapiro2003stochastic} and robust optimization \cite{ben-tal2009robust} offer alternatives, each with its challenges, such as scalability and too conservative solutions. Moreover, modeling realistic repair possibilities exactly is not always possible in such two-stage optimization approaches.  

Decision-focused learning (DFL), extensively reviewed by \cite{mandi2023decisionfocused}, introduces a novel paradigm for stochastic optimization. This approach embeds an optimization model, like Constraint Programming (CP), in a training procedure to minimize a regret loss \cite{SPO}. Challenges arise in backpropagation through combinatorial optimization problems, where solutions may change discontinuously. Recent research, including the score-function approach by \cite{silvestri2023score}, shows promising directions for handling uncertainty in constraints. Both exploring DFL with uncertainty in constraints, and analyzing the applicability of the score-function method are highlighted as valuable directions for further research \cite{mandi2023decisionfocused}.

In this research, we explore various scenario-based approaches for stochastic scheduling. The contribution is threefold: 1) we apply DFL for the first time to a repairable stochastic scheduling problem with stochastic processing times, 2) we demonstrate how an existing DFL technique that uses stochastic smoothing can be used to serve a stochastic scheduling problem where historical realizations of processing times are used, and 3) we include an extensive experimental evaluation in which we assess differences in performance between deterministic, stochastic programming, and a DFL approach.

\section{Scheduling with repair}\label{sec:sched_repair}
We illustrate the effect of uncertainty in constraints with an example of scheduling two tasks on a single machine, where the average task lengths are $\bar{y}_1=4$, and $\bar{y}_2=5$. The machine is not available from $t=5$ to $t=10$. The task is to minimize makespan. Using the mean values, the optimal decision is to schedule first task~2, and then task~1, which gives us a makespan of $14$ (see Figure$~$\ref{fig:subfig1}), while scheduling first task~1, and then task~2 results in a makespan of $15$. 

Now suppose that task~1 is deterministic, and task~2 is stochastic, following the discrete uniform distribution $y_2 \sim U(\{3,4,5,6,7\})$. It still holds that the expected task lengths are $\bar{y}_1=4$, and $\bar{y}_2=5$. When we have $y_2=6$ and we schedule task~2 first, the effect of the repair strategy can be seen in Figure$~$\ref{fig:subfig2} and leads to a makespan of 20. Considering this repair, we can compute the expected values of the two alternative decisions and find that $\mathbb{E}$[first task 1, then task 2] = $15$ and $\mathbb{E}$[first task 2, then task 1] = $16.6$. So, considering the underlying distributions, it is better to first schedule  task~1, instead of task~2. We observe that just using the expected values to come to a decision is not always a good idea when processing times are uncertain. 

\begin{figure}[bt]
\begin{minipage}{0.45\linewidth}
\centering
\begin{tikzpicture}[scale=0.48, font=\fontsize{7}{10}\selectfont]
   \def\taskAstart{0}
   \def\taskAend{2.5}
   \def\taskAlabel{task 2}
   \def\taskBstart{2.5}
   \def\taskBend{5}
   \def\taskBlabel{maint.}
   \def\taskCstart{5}
   \def\taskCend{7}
   \def\taskClabel{task 1}
   
   \draw[fill=red!30] (\taskAstart, 0.5) rectangle (\taskAend, 1);
   \draw[fill=black!30] (\taskBstart, 0.5) rectangle (\taskBend, 1);
   \draw[fill=blue!30] (\taskCstart, 0.5) rectangle (\taskCend, 1);
   
   \node[above] at (\taskAstart + 1 , 0.4) {\taskAlabel};
   \node[above] at (\taskBstart + 1.1, 0.4) {\taskBlabel};
   \node[above] at (\taskCstart + 1, 0.4) {\taskClabel};
 
   \node[below] at (\taskAstart , 0) {0};
   \node[below] at (\taskBstart , 0) {5};
   \node[below] at (\taskCstart , 0) {10};

   \node[below] at (\taskAend , 0) {5};
   \node[below] at (\taskBend , 0) {10};
   \node[below] at (\taskCend , 0) {14};
   \draw[->] (0, 0) -- (11, 0);
\end{tikzpicture}
\caption{Deterministic opt. schedule}\label{fig:subfig1}
\end{minipage}%
\hfill
\begin{minipage}{0.45\linewidth}
\centering
\begin{tikzpicture}[scale=0.48, font=\fontsize{7}{10}\selectfont]
   \def\taskAstart{0}
   \def\taskAend{2.5}
   \def\taskAlabel{fail}
   \def\taskBstart{2.5}
   \def\taskBend{5}
   \def\taskBlabel{maint.}
   \def\taskCstart{5}
   \def\taskCend{8}
   \def\taskClabel{task 2}
   \def\taskDstart{8}
   \def\taskDend{10}
   \def\taskDlabel{task 1}
   
   \draw[fill=red!30] (\taskAstart, 0.5) rectangle (\taskAend, 1);
   \draw[fill=black!30] (\taskBstart, 0.5) rectangle (\taskBend, 1);
   \draw[fill=red!30] (\taskCstart, 0.5) rectangle (\taskCend, 1);
   \draw[fill=blue!30] (\taskDstart, 0.5) rectangle (\taskDend, 1);
   
   \node[above] at (\taskAstart + 1 , 0.4) {\taskAlabel};
   \node[above] at (\taskBstart + 1.1, 0.4) {\taskBlabel};
   \node[above] at (\taskCstart + 1, 0.4) {\taskClabel};
   \node[above] at (\taskDstart + 1, 0.4) {\taskDlabel};
 
   \node[below] at (\taskAstart , 0) {0};
   \node[below] at (\taskBstart , 0) {5};
   \node[below] at (\taskCstart , 0) {10};

   \node[below] at (\taskAend , 0) {5};
   \node[below] at (\taskBend , 0) {10};
   \node[below] at (\taskCend , 0) {16};
   \node[below] at (\taskDend , 0) {20};
   \draw[->] (0, 0) -- (11, 0);
\end{tikzpicture}
\caption{Repair action when $y_2=6$}\label{fig:subfig2}
\end{minipage}
\end{figure}
 \section{Decision-focused learning}\label{sec:dfl}
 \textbf{Problem setting.} The goal is to optimize an optimization (e.g. scheduling) problem  $z^*(y) = \argmin_z f(z,y) \ s.t. \ z \in C(y, z)$, where $f(z,y)$ is the objective function given parameters $y$ and decision $z$ and the constraint set $C(y,z)$. However, the parameters (e.g. processing times) $y$ are unknown at the time of solving. We are given a data set $\mathcal{D}=\{y_i\}_{i=1}^n$ with historical data on $y$. A common approach is to use the sample averages of $\bar{y}$ to solve the deterministic model and obtain $z^*(\bar{y})$ (which possibly requires reparations when the true values become known).  Alternatively, we could take inspiration from the literature on DFL. 
 
  \textbf{Decision-focused learning.} The idea is to predict the unknowns $\widehat{y} = h_{\theta}(\mathcal{D})$ based on the data such that the task loss is minimized. Since the unknown parameters occur in the constraints, predicted decisions must sometimes be corrected using a repair function (such as illustrated in Section \ref{sec:sched_repair}). A common task loss for problems with unknown parameters in the constraints is the so-called post-hoc regret $PRegret$ loss \cite{hu2023branch}, defined as:
\begin{equation}
PRegret(\widehat{y}, y) = f(z_{corr}(\widehat{y}, y), y) - f(z^*(y), y)+ pen(z^*(\widehat{y}), z_{corr}(\widehat{y}, y)),
\end{equation}
where \(y\) are the true coefficient values, \(\widehat{y}\) are the predicted values, \(z^*(\widehat{y})\) is the decision based on predicted values, and \(z^*(y)\) is the optimal decision with perfect information such as defined by \cite{bertsimas2018predictive}. Then, we have \(f(z^*(\widehat{y}),y)\), which are the costs for predicted decisions, and \(f(z^*(y),y)\), which are true optimal costs. Due to uncertain parameters in the constraints, predicted decisions must sometimes be corrected using a repair function such that $z^*(\widehat{y}) \rightarrow z_{corr}(\widehat{y}, y)$. How this reparation is penalized is reflected in $pen(z^*(\widehat{y}), z_{corr}(\widehat{y}, y))$.

  \textbf{Zero-gradient problem.} DFL procedures minimize the post-hoc regret loss by gradient-based optimization with respect to $\theta$ to optimize the prediction $\widehat{y}=h_{\theta}(\mathcal{D})$. However, this loss gives a zero-gradient problem because a combinatorial optimization solver is embedded in the loss computation \cite{SPO}, which is the $\frac{\delta z_{corr}(\widehat{y}, y)}{\delta \widehat{y}}$ term in \eqref{eq:grad}. 
\begin{equation}
\frac{\delta PRegret(\widehat{y}, y) }{\delta \theta} = \frac{\delta PRegret(z_{corr}(\widehat{y}, y), y) }{\delta z_{corr}(\widehat{y},y)} \frac{\delta z_{corr}(\widehat{y}, y)}{\delta \widehat{y}} \frac{\delta \widehat{y}}{\delta \theta} \label{eq:grad}
\end{equation}

\textbf{Stochastic smoothing.} A novel approach by Silvestri et al. \cite{silvestri2023score} shows that this zero-gradient problem can be solved with a stochastic smoothing trick. The crux is to use a stochastic estimator $\widehat{y} \sim p_{\theta}(y)$ (where $p_{\theta}(y)$ is a parameterized distribution) instead of the point estimator $\widehat{y}=h_{\theta}(\mathcal{D})$. Using a stochastic estimator makes the loss function an expectation, for which the gradient can be approximated with the score-function gradient estimator (also known as likelihood ratio gradient estimator \cite{glynn}) that uses: 
\begin{equation}
\nabla_{\theta} \mathbb{E}_{\widehat{y} \sim p_{\theta}(y)} [PRegret(\widehat{y},y)] 
= \displaystyle \mathbb{E}_{\widehat{y} \sim p_{\theta}(y)} [PRegret(\widehat{y},y) \nabla_{\theta} \log(p_{\theta}(\widehat{y}))] \label{eq:grad_sf}
\end{equation}
for which the derivation can be found in \cite{silvestri2023score}. The most important assumption on $p_{\theta}(y)$ is that the probability density function must be differentiable with respect to $\theta$. The right-hand side can be approximated with a Monte-Carlo method~\cite{mohamed2020mcge}. This score-function gradient estimation approach is also the foundation of the REINFORCE algorithm \cite{williams1992reinforce}, and various other reinforcement learning algorithms \cite{sutton}. How we exactly apply these techniques to our stochastic scheduling problem is explained in the next section.
\newpage
\section{From scenarios to schedules} 
\begin{figure}[h]
    \centering
\includegraphics[width=0.45\textwidth]{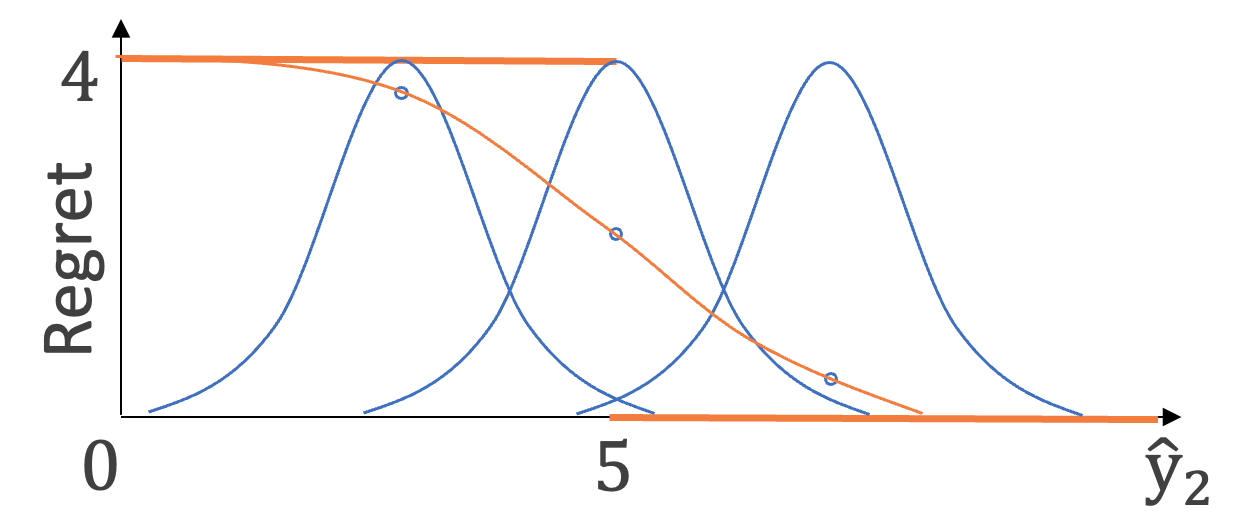}
\caption{Smoothing}\label{fig:smoothing}  \end{figure}

\begin{figure}[h]
\centering
\subfigure{\includegraphics[width=0.4\textwidth]{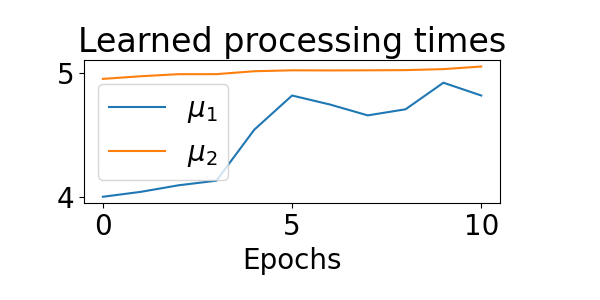}}
\subfigure{\includegraphics[width=0.4\textwidth]{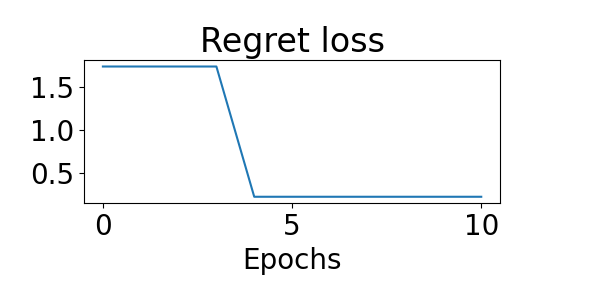}}
\vspace{-0.5cm}\caption{Training curves}\label{fig:toy}
  \end{figure}

\textbf{Algorithm:} We adapt DFL to align with our scheduling problem in Algorithm~\ref{alg:dfl}. The data $\mathcal{D}=\{y_i\}^n_{i=1}$ comprises historical examples of processing times $y$. We aim to learn which predictor $\widehat{y}$ minimizes the post-hoc regret. For gradient computation, we use a stochastic estimator parameterized by $\theta$, for which a common choice is the Normal distribution \cite{sutton}. During training, we sample $\widehat{y} \sim \mathcal{N}(\mu=\theta_{\mu} \cdot \bar{y}, \sigma=\theta_{\sigma} \cdot \bar{\sigma})$, where both $\mu$ and $\sigma$ are trainable, and initially set to the sample average $\bar{y}$ and sample standard deviation $\bar{\sigma}$.  In each training step, we sample a point $y_i$ and a prediction $\widehat{y}$, compute schedule $z^*(\widehat{y})$, and update $\theta$ using the score-function gradient estimator that is provided in equation \eqref{eq:grad_sf}. After training, the stochastic estimator is treated as a point estimator by using $\widehat{y} = \mu$. Note that the distribution is only needed during training for gradient computation on the regret loss. 
 \begin{algorithm}[H]
      \caption{DFL} 
      \label{alg:dfl}
     \begin{algorithmic}
  \REQUIRE $\mathcal{D}_{train}=\{y_i\}^{n_{train}}_{i=1}$, $\mathcal{D}_{test}=\{y_i\}^{n_{test}}_{i=1}$
  \STATE {Initialize $\widehat{y} \sim p_{\theta}(\widehat{y})$  such that $\widehat{y} \sim \mathcal{N}(\mu=\theta_{\mu} \cdot \bar{y}, \sigma=\theta_{\sigma} \cdot \bar{\sigma})$}
  \FOR{each epoch}
  \FOR{each batch in $\mathcal{D}_{train}$}
  \FOR{each instance $(y_i, z^*(y_i)$) in batch}
  \STATE {Sample $\widehat{y}$ from $p_{\theta}(\widehat{y})$}
  \STATE {Pass $\widehat{y}$ to solver 
 to get schedule}
  \STATE {Compute post-hoc regret$(\widehat{y}, y_i)$}
  \ENDFOR 
  \STATE {Update $\theta$ with score-function:}
  \STATE {$\theta = \theta - \text{lr} \cdot \nabla_{\theta} PRegret(\widehat{y},y_i) \nabla_{\theta} \log(p_{\theta}(\widehat{y}))$ }
  \ENDFOR
  \ENDFOR
  \STATE Pass $\widehat{y}=\mu$ to solver to get schedule
  \STATE Evaluate post-hoc regret on  $\mathcal{D}_{test}$
  \end{algorithmic}
    \end{algorithm}
\textbf{Example:} We explain the zero-gradient problem and smoothing technique with our example from Section \ref{sec:sched_repair}. Suppose $(y_1,y_2)=(4,6)$, but $y_2$ is unknown. Figure \ref{fig:smoothing} shows how $\widehat{y}_2$ affects the regret, which is the line with a discontinuity  at $\widehat{y}_2=5$, indicating a jump in scheduling priority. The blue curves show different stochastic estimators, and the small circles the expected regret values when we sample $\widehat{y}_2$ from each distribution. The line through the circles represents the smoothened expected regret. We assess the applicability of Algorithm~\ref{alg:dfl} using this example. The training data has an underlying distribution with $y_1 = 4$ and $y_2 \sim U({3,4,5,6,7})$. We expect the algorithm to find scaling $\theta_2 > 1$ to prioritize scheduling task~1. A small experiment confirms that regret drops when $\mu_1$ is above five as anticipated, see Figure \ref{fig:toy}.

\section{Experimental evaluation}
To understand the potential of DFL, we compare performance to deterministic and stochastic programming formulations (Section~\ref{sec:baseline}). We hypothesize that both stochastic programming and DFL outperform the more naive deterministic approach. Furthermore, we explore when DFL or stochastic programming performs better; we expect that DFL has better scalability to larger instances.

\subsection{Problem instances and evaluation}\label{sec:data}
We study a variant of the \textbf{Stochastic Resource Constraint Project Scheduling Problem} (RCPSP). We use two subsets of the PSPLib instances (being j301$\_$1 to j303$\_$10, and j901$\_$1 -to j903$\_$10) \cite{Kolisch1996PSPLIBA}. Furthermore, we use two sets of industry-inspired problem instances (small and large instances\footnote{25 instances with 40 to 480 tasks, 13 resource groups with different capacities.}), related to the factory of our industrial partner DSM-Firmenich. The data describing these instances is provided in our repository \cite{repo}. 
Originally deterministic, these instances are transformed into stochastic versions by sampling task durations from Normal distributions with mean $d_j$ and standard deviation $\sqrt{d_j}$, where $d_j$ is the deterministic processing time of task $j$ from the original instance. We define a scenario as a processing time vector realization for one problem instance. For each instance, three datasets are created: one with 100 training scenarios, another with 50 for validation and tuning, and a final set of 50 for evaluation.

 The \textbf{evaluation} approach evaluates first-stage start times $z_j$ decisions based on the schedule makespan. Tasks unable to start due to resource constraints or precedence relations undergo a repair policy with (multiple) one-unit time postponements resulting in corrected start times \(z_j^{corr}\). A penalty function measures the sum of start time deviations for all tasks \(j\): \begin{equation} \text{pen}(z^*(\widehat{y}), z^*_{corr}(\widehat{y}, y)) = \rho \cdot \displaystyle \sum_{j \in J} z_j^{corr}(\widehat{y}, y) - z_j(\widehat{y}).\end{equation} Here, \(\rho\) is the penalty coefficient, which we vary across experiments. Evaluation is conducted using the SimPy discrete-event simulation Python package \cite{simpy}.

\subsection{Baseline methods}\label{sec:baseline}
We study problem cases where historical realizations of stochastic processing times are available (without feature data). This section describes two other scenario-based methods that are included in the experiments.

\textbf{Deterministic approach.} This is a simple baseline, where we compute scenario averages of the unknown optimization coefficients. The deterministic constraint programming (CP) model that uses these averages uses the following nomenclature: $J$: set of all tasks, $R$: set of all resources, $S_j$: set of successors of task $j$, $j$: subscript for tasks, $r$: subscript for resources,  \textit{parameters:} $y_j$: processing time of task $j$, $r_{r,j}$: resource requirement for task $j$, $b_r$: max capacity of resource $r$, $\text{minLag}_{j, i}$: min.\ difference between start times of tasks $j$ and $i$, if $i$ is a successor of $j$ and \textit{decision variables:} $x_j$: interval length for task $j$. The CP model is: \begin{subequations}
\begin{align}
\text { Minimize } Makespan  & \quad \quad \text{ s.t.} \\
 \mathrm{Max(end\_of(x_j))} \leq Makespan & \quad \quad j \in J \label{cons:makespan} \\
 \mathrm{startOf(x_i)} \geq \mathrm{endOf(x_j)}; & \quad \quad \forall j \in J \ \forall i \in S_j \nonumber \quad or \\
\mathrm{startOf(x_i)} \geq \mathrm{minLag_{j,i}} + \mathrm{startOf(x_j)}; & \quad \quad \forall j \in J \ \forall i \in S_j \label{cons:precedence} \\
 \sum_{j \in J} \mathrm{Pulse(x_j, r_{r,j}) \leq b_r}  & \quad \quad \forall r \in  R  \label{cons:resource}  \\
x_j: \mathrm{IntervalVar(J, y_j)} & \quad \quad \forall j \in J
\end{align}
\end{subequations} In this model, \eqref{cons:makespan} defines the makespan which should be larger than the finish time of all tasks, and \eqref{cons:precedence} enforces precedence constraints between two tasks, where the $\mathrm{minLag_{a,b}}$ is the minimal time difference needed between task $a$ and $b$ which is used for the industry instances. The CP pulse constraint \eqref{cons:resource} models shared resource usage \cite{cplex2009v12}.

\textbf{Stochastic programming.} The second baseline comprises a scenario-based stochastic programming formulation (again CP). The repair action is added to the stochastic model which comprises the possibility to postpone activities, together with a penalty term for the deviations from the earliest-start-time decision that is included in the objective. Note that we use the same nomenclature as for the deterministic model, but we introduce the notion of scenarios $\omega \in \Omega$, and the first-stage earliest-start-time decision variable $z_j \ \forall j \in J$.

\begin{subequations}
\begin{align}
   \text{ Min} \frac{1}{|\Omega|} \sum_{\omega \in \Omega} Makespan({\omega}) +  \rho \cdot \sum_{\omega \in \Omega} \sum_{j}  
 &\mathrm{startOf(x_j(\omega))} - z_j \quad  \text{s.t} \\
 \mathrm{Max_j(end\_of(x_j(\omega)))} \leq Makespan({\omega}) & \quad \quad \forall \omega \in \Omega \\
\mathrm{startOf(x_i(\omega))} \geq \mathrm{endOf(x_j(\omega))}; & \quad \quad \forall j \in J \ \forall i \in S_j \nonumber \quad or \\
\mathrm{startOf(x_i(\omega))} \geq \mathrm{minLag_{j,i}} + \mathrm{startOf(x_j(\omega))}; & \quad \quad \forall j \in J \ \forall i \in S_j \quad \forall \omega \in \Omega \\
 \sum_{j \in J} \mathrm{Pulse(x_j(\omega), r_{r,j}) \leq b_r}  & \quad \quad \forall r \in  R \quad \forall \omega \in \Omega \\
x_j(\omega): \mathrm{IntervalVar(J, y_j(\omega))} & \quad \quad \forall j \in J \quad \forall \omega \in \Omega \\
 z_j \leq  \mathrm{startOf(x_j(\omega))} & \quad \quad \forall j \in J \quad \forall \omega \in \Omega 
\end{align}
\end{subequations}

\subsection{Results}
 \begin{figure}[h]
     \subfigure[PSPlib j30]{\includegraphics[width=0.2\textwidth]{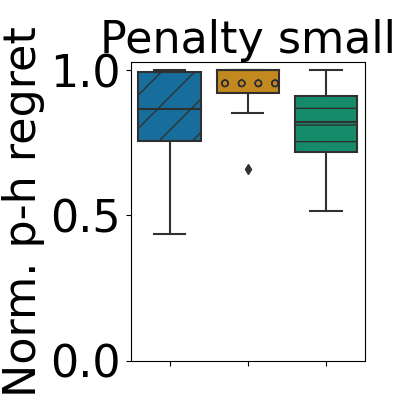}}
         \subfigure[PSPlib j30]{\includegraphics[width=0.2\textwidth]{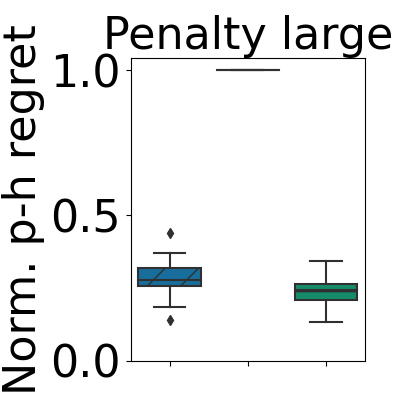}}
        \subfigure[PSPlib j90]{\includegraphics[width=0.2\textwidth]{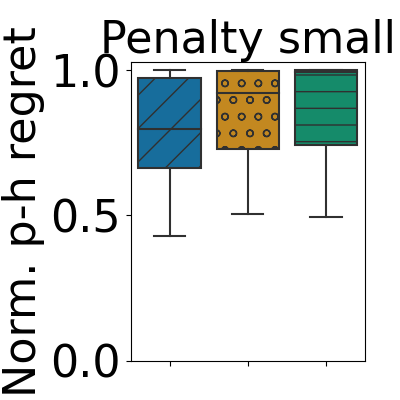}}
         \subfigure[PSPlib j90]{\includegraphics[width=0.2\textwidth]{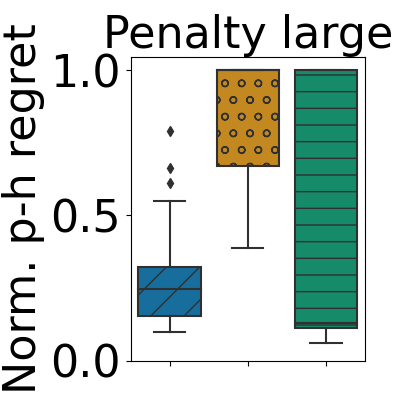}}
\subfigure{\includegraphics[width=0.15\textwidth]{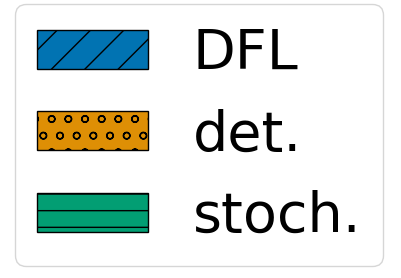}}
    \subfigure[Small indus.]{\includegraphics[width=0.2\textwidth]{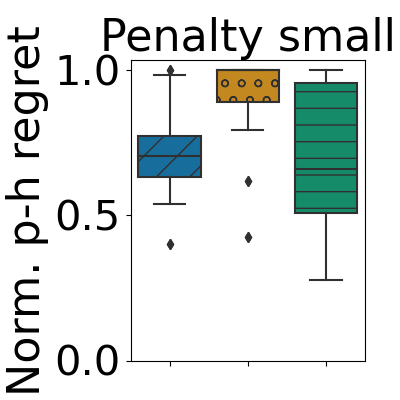}}
  \subfigure[Small indus.]{\includegraphics[width=0.2\textwidth]{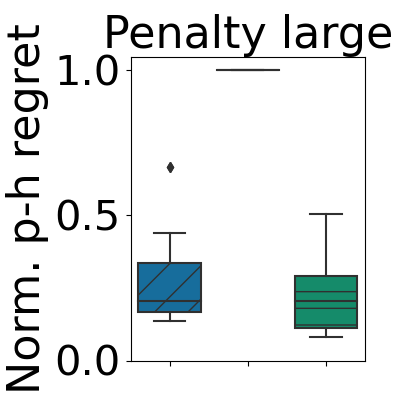}}
    \subfigure[Large indus.]{\includegraphics[width=0.2\textwidth]{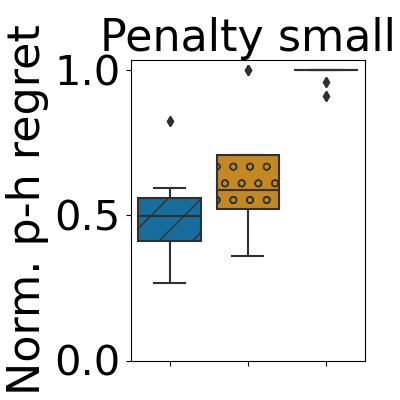}}
  \subfigure[Large indus.]{\includegraphics[width=0.2\textwidth]{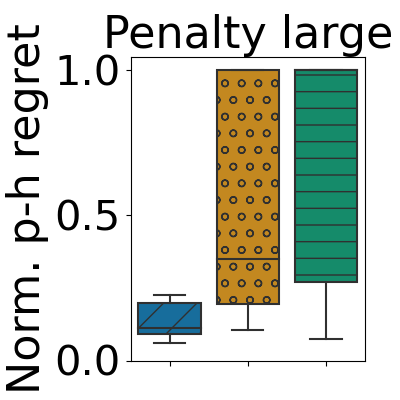}}
\subfigure{\includegraphics[width=0.15\textwidth]{images/results/legend_v2.png}}
           \caption{Normalized post-hoc regret per instance set - penalty setting (smaller regret is better). The box spans from the 25th to the 75th percentile, visualizing the median and interquartile range.} \label{fig:boxplot}
\end{figure}
All experiments are done on a virtual server that uses an Intel(R) Xeon(R) Gold 6148 CPU with two 2.39 GHz processors, and 16.0 GB RAM. All CP models are solved with single thread IBM CP solver \cite{cplex2009v12}. The runtime limits are set per problem size (max.\ 60min.) and provided in the README of the repository \cite{repo}, together with the tuned hyperparameters\footnote{Such as the number of scenarios.} for deterministic, DFL, and stochastic. Each boxplot in Figure$~$\ref{fig:boxplot} presents the results for the three methods on a single combination of instance set and penalty setting. The y-axis shows the distribution of the normalized post-hoc regret among the test instances of that specific set. We use $\rho=\frac{1}{size}$ (small), where $size$ indicates the number of tasks, and $\rho=1$ (large). We tested the significance of the performance difference of the different algorithms for each setup (a-i) using a paired t-test with $\alpha=0.05$, and the p-values are included in the repository \cite{repo}.

On smaller instances with small penalties, both DFL and stochastic methods perform well, with no significant difference. For the large penalty, stochastic tends to outperform both deterministic and DFL methods significantly in PSPlib \(j30\) instances. Notably, under \(\rho=1\) for \(j30\) instances, no repairs were needed across all instances, emphasizing the robust performance of stochastic when the instances are small enough. For larger instances like PSPlib \(j90\), DFL becomes better, even significantly for the small penalty. For the large penalty, there is still a subset of the instances for which stochastic finds very robust solutions that do not need repairs, but because for some of the instances the stochastic model performs much worse than DFL (which is also visible in Figure \ref{fig:boxplot}d), we observe no significant difference between DFL and stochastic looking at all \(j90\) instances. We see a somewhat similar pattern in the industrial instances, where again stochastic is most advantageous for the smaller instances and with a high penalty, although not significantly better than DFL. For the larger instances, stochastic performs even worse, especially for the small penalty and DFL is significantly better. We investigated optimality gaps of the outputs of the stochastic model and found that even with a time limit of three hours the gaps are on average approximately around $30\%$, with outliers of more than 90 $\%$ (for the largest industry instances) which shows the scalability issue of stochastic programming.

\section{Related Work}
Previous studies \cite{berthet2020learning,SPO,mandi2023decisionfocused} focused mainly on comparing prediction-focused versus DFL approaches for problems with uncertainty in (linear) objectives. The knapsack problem is the most prominent \cite{Demirovic2019AnII,mandi2020predictandoptimize}. As far as we know, the set-up without feature data is not studied in earlier work \cite{mandi2023decisionfocused}. However, the crux of our problem setting is that uncertain parameters occur in constraints, which can lead to infeasibilities. Hu et al.~\cite{hu2022predictoptimize} were the first who introduced a post-hoc regret that penalizes for infeasibilities.  The methods introduced in their work rely on specific conditions, such as being recursively and iteratively solvable \cite{hu2023branch,hu2022predictoptimize}. This work applies a DFL approach \cite{silvestri2023score} to a pure stochastic repairable scheduling problem, for which both the repairable scheduling setting and the context without features are novel application domains for DFL. It is important to highlight that before this study we did not know if DFL could work for repairable scheduling.

\section{To conclude, and continue}
This study explores a novel application of DFL to stochastic resource-constrained scheduling with repairs, where uncertainty is in the constraints, and the derivative is not smooth by itself. Results indicate that stochastic programming is dominant when it can find the optimal solution, most prominently when the penalty factor is high and the instances are small enough to find robust solutions that do not need reparation. In contrast, we have shown that DFL scales better and is a promising alternative to stochastic programming, even in this pure stochastic scheduling setup. Furthermore, we highlight the potential of DFL because of its flexibility across various settings with different repair strategies, providing a distinct advantage over stochastic programming, in which modeling the exact repair functions is not always possible. We hypothesize that in a setting with features related to stochastic processing times the benefits of DFL for stochastic scheduling are further enhanced, such as shown in earlier research with uncertainty in a (linear) objective \cite{mandi2023decisionfocused}. Further interesting directions are investigating alternative gradient estimators or reinforcement learning-inspired algorithms.

\subsubsection*{Acknowledgements}
We acknowledge Mattia Silvestri and Michele Lombardi whose valuable insights improved the quality of this work. We are grateful that Léon Planken was available for questions related to the simulator. This work is supported by the AI4b.io program, a collaboration between TU Delft and dsm-firmenich, and is fully funded by dsm-firmenich and the RVO (Rijksdienst voor Ondernemend Nederland). 

\subsubsection*{Disclosure of Interests}
The authors have no competing interests to declare that are relevant to the content of this article.
\bibliographystyle{alpha}
\bibliography{template} 

\end{document}